%% file: Main.tex
\documentclass[lettersize,journal]{IEEEtran}
\usepackage{amsmath,amsfonts}
\usepackage{algorithmic}
\usepackage{algorithm}
\usepackage{array}
\usepackage[caption=false,font=normalsize,labelfont=sf,textfont=sf]{subfig}
\usepackage{textcomp}
\usepackage{stfloats}
\usepackage{url}
\usepackage{verbatim}
\usepackage{graphicx}
\usepackage{cite}

\usepackage{multirow} 
\begin{document}

\title{Bayesian Optimization Framework for Efficient Fleet Design in Autonomous Multi-Robot Exploration}

\author{David Molina Concha, Jiping Li,   Haoran Yin,  Kyeonghyeon Park, Hyun-Rok Lee, Taesik Lee,  Dhruv Sirohi,  Chi-Guhn Lee


\thanks{This work was supported in part by the International Doctoral Cluster Program of the University of Toronto,  \#DSF20-01; and
in part by the National Agency for Research and Development of Chile, program Beca Chile Doctorado Extranjero \#72200533.  }
\thanks{David Molina Concha and Chi-Guhn Lee (Corresponding author) are with the Dynamic Optimization and Reinforcement Learning Laboratory of the Department of Mechanical \& Industrial Engineering,  Jiping Li and Haoran Yin are with the Department of  Electrical \& Computer Engineering,  Dhruv Sirohi is with the Division of Engineering Science of the Faculty of Applied Science and Engineering, University of Toronto, Toronto, ON M5S3G8, Canada. (e-mail david.molina@mail.utoronto.ca; cglee@mie.utoronto.ca; jiping.li@mail.utoronto.ca; haoran.yin@mail.utoronto.ca; dhruv.sirohi@mail.utoronto.ca). 

Kyeonghyeon Park and  Taesik Lee are with the Complex System Design Laboratory of the Department of Industrial and Systems Engineering, Korea Advanced Institute of Science and Technology, Daejeon, 34141, Korea.(e-mail kyeonghyeon.park@kaist.ac.kr; taesik.lee@kaist.ac.kr).

Hyun-Rok Lee is with the  Department of Industrial Engineering, Inha University, Incheon,  22212, Korea. (e-mail hyunrok.lee@inha.ac.kr).}}


\maketitle

\begin{abstract}

This study addresses the challenge of fleet design optimization in the context of heterogeneous multi-robot fleets, aiming to obtain feasible designs that balance performance and costs. In the domain of autonomous multi-robot exploration, reinforcement learning agents play a central role, offering adaptability to complex terrains and facilitating collaboration among robots. However,  modifying the fleet composition results in changes in the learned behavior, and training multi-robot systems using multi-agent reinforcement learning is expensive. Therefore, an exhaustive evaluation of each potential fleet design is infeasible. To tackle these hurdles, we introduce Bayesian Optimization for Fleet Design (BOFD), a framework leveraging multi-objective Bayesian Optimization to explore fleets on the Pareto front of performance and cost while accounting for uncertainty in the design space. Moreover, we establish a sub-linear bound for cumulative regret, supporting BOFD's robustness and efficacy. Extensive benchmark experiments in synthetic and simulated environments demonstrate the superiority of our framework over state-of-the-art methods, achieving efficient fleet designs with minimal fleet evaluations.

\end{abstract}

\begin{IEEEkeywords}
Design optimization, Multi-robot systems, Autonomous agents, Deep reinforcement learning.
\end{IEEEkeywords}

\section{Introduction}
\IEEEPARstart{T}{he} design and optimization of heterogeneous multi-robot systems for various applications, such as environmental monitoring, search and rescue operations, and industrial inspection, have been the focus of extensive research recently due to the advantages over homogeneous fleets \cite{roldan2021reviewmultirobot}. With the increasing complexity of tasks and the imperative need for efficient resource allocation, researchers have honed their focus on enhancing coordination among robots. This enhancement takes the form of developing goal assignment mechanisms \cite{gautieretal2023mrtask, baietal2023auctiontask, al2024taskreall} and reliable communication mechanisms \cite{Mannucci2019unreliablecommunication, Goel2020aircommunication, Bruggemann2009signalmulti}. Furthermore, the escalating complexity of missions has given rise to the demand for larger fleet sizes and greater heterogeneity in robot capabilities \cite{gerkey2004formal}. Despite these challenges, determining the optimal composition of a heterogeneous fleet for efficient exploration has seen partial efforts within the context of autonomous robots.

The fleet design problem involves a delicate balance between performance and costs \cite{cai2021energyfleet}. The selection of hardware components within a multi-robot fleet significantly impacts its overall performance. For example, a robot type with an extended sensor should be able to explore better than another robot type with a shorter sensor range. Opting for full-featured robots might seem attractive as a means to maximize fleet performance. However, it's essential to acknowledge that such component choices also come with substantial budget considerations. Take into account the impressive capabilities of a full-featured ground robot like Boston Dynamics' Spot, which comes at a significant cost of US \$74,000 ~\cite{turtleb32023price}. In contrast, a more budget-friendly, yet less feature-rich option like the Turtlebot3 Burger can be acquired for a substantially lower price, around US \$600 ~\cite{spot2023price}. The balance between cost and performance is a pivotal factor in designing an efficient fleet composition to effectively achieve the exploration task.

In the domain of multi-robot exploration, autonomous robots are frequently trained as reinforcement learning (RL) agents. This choice is driven by several reasons. Firstly, RL empowers robots to adapt and learn from their interactions with the environment and the feedback from their actions. This adaptability allows them to navigate and explore unfamiliar or complex terrains, resulting in heightened efficiency and effectiveness in exploration tasks \cite{miller2020mineexploration}. Secondly, RL facilitates coordination and cooperation among multiple robots working as a team. By training robots as RL agents, they can master communication and collaboration, enabling them to share information and coordinate actions to achieve common exploration goals \cite{silva2022comaware}. This teamwork significantly enhances the exploration capabilities of the robot fleet by harnessing collective knowledge and experiences.

The performance of a multi-agent system of RL agents depends on their joint behavior, learned over multiple training episodes. Introducing, removing, or replacing agents in the system results in changes in the learned behavior, necessitating retraining of the entire fleet of agents. Multi-agent reinforcement learning (MARL) algorithms face significant computational challenges due to the exponential increase in state and action space relative to the number of agents \cite{qu2020scalable}. This scalability limitation makes it infeasible to explore the behavior of all possible fleet combinations with multiple heterogeneous agents.

In the context of optimizing expensive-to-evaluate functions, Bayesian Optimization (BO) emerges as a potent tool for addressing such problems \cite{srinivasetal2010lemma51}. BO leverages probabilistic models to guide the search for optimal solutions, effectively reducing the number of expensive function evaluations required. Single-objective BO has recently found application in algorithmic reward design within MARL \cite{mguni2019coordinating, shou2020reward}. This approach is particularly suitable for assessing promising reward parameters in continuous spaces with a limited evaluation budget. However, extending these methods for fleet design requires additional efforts for discrete search space and formulating the problem as single-objective might not capture the trade-off between performance and cost.  

While multi-objective (MO) optimization \cite{Henry2023fleetoptimization} can handle the trade-off between acquisition costs and performance, there exists no prior work addressing algorithmic approaches to solving fleet design in the context of autonomous robots. To address this problem efficiently, we propose Bayesian Optimization for Fleet Design (BOFD), a bi-level optimization framework. The upper-level problem utilizes Multi-Objective Bayesian Optimization (MOBO) to optimize fleet composition. In the lower-level problem, it evaluates new fleet designs by training agents using MARL algorithms. Our contributions are:

\begin{itemize}
    \item We introduce an efficient framework for addressing the fleet design problem with heterogeneous autonomous robots, requiring only a few fleet evaluations.
    \item We establish a sub-linear regret bound with respect to the number of iterations, supporting the effectiveness of our approach.
    \item We showcase the performance of the BOFD framework through comprehensive evaluations in synthetic and simulated environments. Our benchmark studies include comparisons against state-of-the-art methods in both single and multi-objective Bayesian Optimization.
\end{itemize}

\section{Related Work}

The impact of fleet size in exploration tasks is studied in \cite{yan2014team}, showing that increasing fleet size does not always lead to improvement in performance due to the efforts to avoid collision with other robots while exploring the environment.  Their work provided insights into the effects of modifying the number of robots in the system without proposing an algorithmic solution for the fleet design problem. Similarly,  \cite{barrios2014fleetsize} highlights the importance of considering the trade-off between fleet size and efficiency in autonomous car systems without delving into an algorithmic method for fleet design. The work of \cite{cabreramoraetal2014fleetgraph} explores the trade-offs involved in the selection of different fleet sizes, contributing a mathematical analysis of the time gains versus costs. Their discussion is in the context of homogeneous robots using graph search algorithms for exploration. 

In the domain of reward design, \cite{mguni2019coordinating} introduces a bi-level optimization for the reward design problem, optimizing a continuous reward parameter at the upper level using single-objective BO  and solving the underlying Markov Game at the lower level using a MARL algorithm. \cite{shou2020reward} implemented the bi-level framework to design rewards for transportation firms. While it is feasible to extend these frameworks to address the fleet design problem, single-objective Bayesian Optimization is unable to effectively capture the trade-off between cost and performance.
 
The fleet design problem has been modeled as a multi-objective optimization problem in \cite{Henry2023fleetoptimization}, presenting a general method for integrating multi-objective system design and fleet acquisition planning. Their work is focused on the interdependence of both problems via mixed-integer linear programming, without dealing with autonomous agents or expensive-to-evaluate functions. For addressing the fleet design problem efficiently, multi-objective Bayesian optimization methods, such as USeMO \cite{Belakaria2020usemo}, offer valuable approaches. USeMO employs Gaussian processes to model black-box objective functions, constructing a cheap MO problem. It selects solutions from the Pareto front that maximize the volume of uncertainty hyper-rectangles. While USeMO exhibits simplicity and sub-linear regret properties, making it computationally efficient and competitive with state-of-the-art methods, it lacks consideration of inherent characteristics specific to fleet design optimization. Moreover, its effectiveness in the context of fleet design remains untested.

\section{Preliminaries}

\subsection{Multi-robot exploration} \label{subsec:mrexploration}

Multi-robot exploration involves deploying multiple robots to explore and map an environment efficiently. The primary objective in multi-robot exploration is to identify the optimal path that minimizes either the total distance traveled or the time taken to explore the entire environment. In the context of this optimization, the objective function of the exploration problem aims to determine the path $P^*$ that minimizes the overall distance traveled or time spent from the current location of each robot to a specific destination point 
$p$. Assuming a grid-like map, the optimization problem can be formulated as follows:

\begin{equation}
    P^* = \arg \min_{p \in \mathcal{P}} \sum^N_{i=1} \sum^{T_{\text{max}}}_{t=1} g(p^i_{t}),     
\end{equation}

\noindent here, $\mathcal{P}$ denotes the set of all points on the grid map, $T_{\text{max}}$ represents the total time steps of robot $i$, $N$ is the total number of robots in the fleet, $p^i_t$ is the destination point chosen by robot $i$ at time $t$, and $g(p)$ is a function describing the distance or time from robot $i$'s current location to the destination point $p$.

Deploying multiple robots offers advantages such as concurrency, reduced mission time, and the ability to perform tasks more efficiently than single robots \cite{Lee2021multirobotconcu}. However, it presents challenges related to coordination, merging information obtained by several robots, and dealing with limited communication \cite{bietal2023aimexp}. To address these challenges, RL has been applied to enable robots to learn and adapt their exploration strategies.

\subsection{Partially Observable Markov Games}

The application of RL in multi-robot exploration is significant as it allows robots to optimize their exploration strategies based on the received rewards. This is particularly relevant in scenarios where robots need to coordinate in challenging terrains and large-scale geometries, as it enables them to adapt their behavior to maximize the exploration efficiency \cite{Kulkarni2022multirobotairground}. Furthermore, the introduction of RL aligns with the need for robots to learn and adapt to the environment during the exploration process.

The multi-robot exploration problem is traditionally modeled as a Partially Observable Markov Game (POMG), which is an extension of a Partially Observable Markov Decision Process (POMDP) ~\cite{pack1998POMG} to the multi-agent scenario. Specifically, the POMG is defined by the tuple:
\begin{align}
    <N, S, O, \{\sigma_i\}, A, \{R_i\}, T, \gamma>,
\end{align}

\noindent where $N$ represents the set of agents with $i \in N$,  $S$ is the state space shared by all agents, and $O \equiv O_1 \times \ldots \times O_N$ and $A \equiv A_1 \times \ldots \times A_N$ are the observation and action spaces, with $O_i \subseteq S$ for all $i \in N$. Moreover, $\{\sigma_i\}$ and $\{R_i\}$ represent the observation and reward functions for all $i \in N$, where $\sigma_i : S \rightarrow O_i$ and $R_i : S \times A \times S \rightarrow \mathbb{R}$, respectively. The stochastic state transition model $T$ is given by $T: S \times A \times S \rightarrow [0, 1]$. Finally, $\gamma$ represents the discount factor.

At each time-step $t$, agent $i$ receives an observation $o_{t,i} = \sigma_i(s_t) \in O_i$ which corresponds to a portion of the global state $s_t \in S$. Each agent employs a stochastic policy $\pi_i$ to select an action $a_{t,i} \sim \pi_i(\cdot | o_{t,i})$. The agents' joint actions $\mathbf{a}_t = (a_{t1}, \ldots, a_{tN}) \in A$ and the current state $s_t$ are used in the transition model to generate the next state $s_{t+1} \sim T(\cdot | s_t, \mathbf{a}_t)$. The obtained reward $r_{t,i} = R_i(\cdot | s_t, \mathbf{a}_t, s_{t+1})$ is then provided to agent $i$.

The primary objective of each agent is to maximize the sum of discounted rewards $V_{i} = \sum_{k=0}^{T} \gamma^k r_{t+k, i}$ over an episode with a finite or potentially infinite horizon \(T\). Each agent seeks to optimize its value function \(V_i(o_{t,i}) = \mathbb{E}_{\pi_i}[\mathcal{V}_{i} | o_{t,i}]\), which represents the expected return starting from observations \(o_{t,i}\) and following policy \(\pi_i\).

Recent work in RL for multi-robot exploration has focused on the use of centralized training for decentralized execution (CTDE) \cite{tan2023macroact}. This approach uses a centralized controller to coordinate the actions of multiple robots and learn a global policy. This centralized training allows for the exchange of information and coordination among the robots, leading to improved exploration efficiency and performance \cite{he2020DMEDRL}. However, during the execution phase, no centralized information is required, and each robot acts autonomously based on its own observations and learned policy. This decentralized execution is desirable in many multi-robot systems as it reduces the reliance on communication and allows for robust and scalable operation.

\subsection{Bayesian optimization}

Evaluating the performance of each fleet design is challenging due to the sheer computation required to obtain $\pi_i$  for all $i \in N$ using MARL. BO is a suitable approach for optimizing single-objective  Markov Games' design with a limited number of evaluations \cite{shou2020reward,mguni2019coordinating}. BO sets a Gaussian process (GP) to model the system's objective $\mathcal{F(\mathbf{n})}$ evaluated at some values of $\mathbf{n}$ as prior. The GP is characterized by the mean $\mu$, which is often set to 0,
and a kernel function $\kappa$ that captures the flexibility and generalization ability of the GP. By using acquisition functions (AF) as the lower confidence bound (LCB), we can select the next design $\mathbf{n}$ to be evaluated by minimizing regret through the optimization process \cite{srinivasetal2010lemma51}. LCB is calculated as follows:
\begin{equation}
\text{LCB}(\mathbf{n}) = \mu^*(\mathbf{n}) - \sqrt{2\text{log}(t^{d/2 + 2} \pi^2/3\delta)}\sigma^*(\mathbf{n}), \label{eq:lcb}
\end{equation}\label{LCB}

\noindent where $\mu^*(\mathbf{n}), \sigma^*(\mathbf{n})$ are the posterior mean and the standard deviation at $\mathbf{n}$, $d$ is the dimension of search of space $\mathbf{n}$, $t$ is the iteration number in BO, and $\delta$ is a parameter.

The extension of BO for multi-objective problems provides powerful techniques to search for optimal trade-offs between conflicting black-box objectives \cite{qing2023robustmobo}. In MOBO, new points are sequentially added using an infill criterion that guides the search towards the Pareto front. The state-of-the-art approach USeMO \cite{Belakaria2020usemo}, generates a posterior distribution by fitting a GP in each of the objectives independently to build a cheap MO problem using the AF, defined as:
\begin{equation}
    N^* \leftarrow min_{\mathbf{n}\in N} (AF_{F_1}(\mathbf{n}),...,AF_{F_k}(\mathbf{n})), 
\end{equation}

\noindent where $N^*$ is the Pareto set , $\mathbf{n}\in N$ is the decision variable, $AF_{F_j}(\mathbf{n})$ is the AF of the objective $F_j$, where $j \in [1,...,k]$. This formulation allows the use of computationally efficient solvers to obtain the Pareto front of solutions, as the GP-AF is given to any point in the search space.


\section{Problem Definition}

In this work, we consider physical heterogeneous fleets ~\cite{bettini2023heterogeneous}, where each type of robot possesses a distinctive combination of hardware components that differs from the other types. We define the POMG to consider multiple types of agents as follows:
\begin{itemize}
    \item $N$ is the set of fleets, with each fleet $\mathbf{n} \in N$ represented as an $M$-dimensional vector. Here, $M$ represents the number of available types of agents and each entry in the vector is the number of agents of a specific type. 
    \item $S$ is the global state of the system, which collects the individual observation of the agents and the environment.  
    \item $o_i$ considers the local information agent $i$ has collected given the characteristic of its sensor. Robots can exchange local information if they are close to each other.
    \item $A$ assumes robots can take actions to move in any of the four cardinal directions and also diagonally to explore the environment. 
    \item $R$ agents receive a reward based on the area explored at every time step and the total distance moved as in \cite{he2020DMEDRL}. The reward function promotes exploration of the whole environment while minimizing the traveled distance. 
\end{itemize}

The fleet design problem can be modeled as a single-objective problem $\min_{\mathbf{n} \in N} \mathcal{F}(\mathbf{n})=P(\mathbf{n})+C(\mathbf{n})$, where $P(\mathbf{n})$ is the fleet performance and $C(\mathbf{n})$ is the acquisition cost for the fleet $\mathbf{n}$.  However, in the context of Bayesian Optimization, conflicting objectives are better captured by formulating the problem as multi-objective \cite{Shu2020newmobo}. Therefore, we model the trade-off between performance and cost as a multi-objective optimization problem, defined as $\min_{\mathbf{n} \in N} [P(\mathbf{n}), C(\mathbf{n})]$. 

As mentioned in Section \ref{subsec:mrexploration}, the performance of the multi-robot system is often measured as the total time to explore the environment. The system performance depends on the joint policy of the robots in the fleet $\boldsymbol{\pi}$. Obtaining $\boldsymbol{\pi}$ is not trivial, as we have to solve the POMG using MARL algorithms, which are computationally expensive in large state and action spaces. As a result, in this work, we assume the fleet performance $P(\mathbf{n})$ as an expensive-to-evaluate function. For the cost function, we follow a linear acquisition cost $C(\mathbf{n})=\mathbf{A}\mathbf{n}$, where $\mathbf{A}$ is a $M$-dimensional column vector containing the acquisition cost for all the $M$ types of exploration robots. This formulation means that the multi-objective problem is a hybrid with one black-box function, $P(\mathbf{n})$, and a known function $C(\mathbf{n})$. We leverage the hybrid objective to propose a tailor-made method for the fleet design problem, isolating the uncertainty from the black-box function and building a cheap MO problem using the AF of $P(\mathbf{n})$.   


\section{Methodology}

We introduce a novel framework, named Bayesian Optimization for Fleet Design (BOFD), tailored to address fleet design challenges through the utilization of Multi-Objective Bayesian Optimization for multi-robot systems. In this section, we start by giving an overview of BOFD. Then, we delve into the specifics of its main components. Afterward, we offer a theoretical analysis of BOFD, focusing on its asymptotic regret bounds.

\subsection{Bayesian optimization for fleet design}

\begin{figure}[!t]
\centering
\includegraphics[width=3.3in]{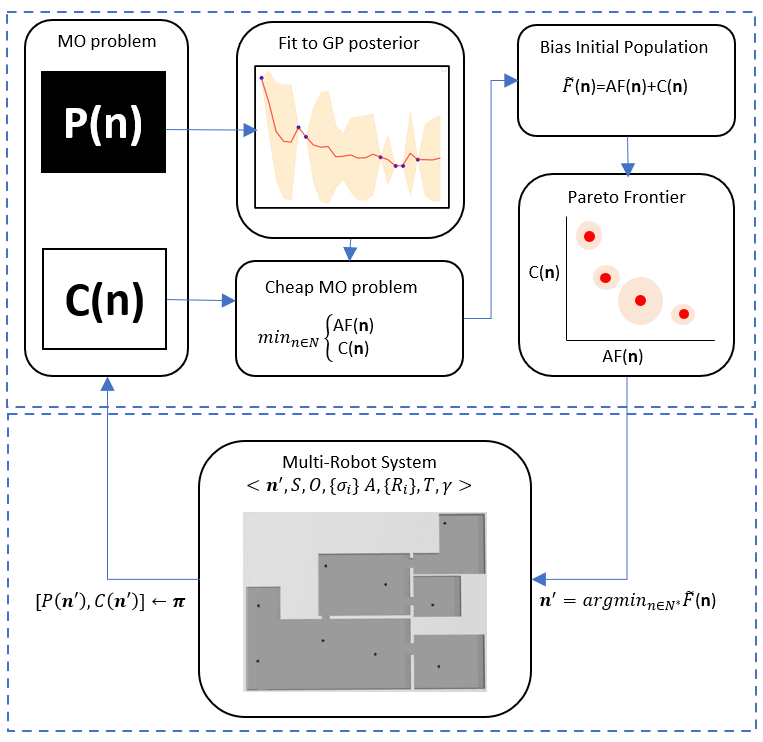}
\caption{BOFD architecture using MO BO to obtain optimal fleet $\mathbf{n}$ at the upper level and the joint policy $\bf{\pi}$  is sought for the given $\mathbf{n}$ at the lower level multi-robot system.}
\label{fig:framework}
\end{figure}

The proposed architecture for BOFD, shown in Figure \ref{fig:framework}, addresses fleet design as a multi-objective problem at the upper level to find promising fleet composition $\mathbf{n}$ and then, at the lower level, we obtain the system performance $P(\mathbf{n})$ by computing agents' joint policy $\boldsymbol{\pi}$ for the given fleet $\mathbf{n}$. 

At the upper level, we set as posterior a GP over the performance of the fleets representing the kernel as the product of $M$ squared exponential (SE) kernels, which works well in discrete search with multiple variables \cite{duvenaud2014kernel}. We build a cheap MO problem as in \cite{Belakaria2020usemo}, using the AF score of the performance and the known cost function. We leverage the cheap MO information to generate a biased initial population score for the genetic algorithm, employing the single-objective formulation $\tilde{F}(\mathbf{n})=AF(\mathbf{n})+C(\mathbf{n})$. This method of initial population assessment promotes fleets with high potential and high uncertainty, where the later is captured by the AF. Once the genetic algorithm obtains the Pareto front of the cheap MO problem, we have to select a solution from the front as a candidate to evaluate its performance, for this purpose, we select the solution that minimizes $\tilde{F}(\mathbf{n})$. The selection rule from the Pareto front prioritizes fleets with high potential and high uncertainty. This strategy ultimately works towards minimizing uncertainty in the predicted MO function.  

Since we present the MO problem as a minimization task, we propose LCB as the AF. LCB is a weighted difference of the expected performance of the fleet captured by $\mu^*(\mathbf{n})$ of the GP, and of the uncertainty captured by the standard deviation of the GP $\sigma^*(\mathbf{n})$. The balance between exploitation and exploration is dynamically adjusted by following the formulation provided in Equation \ref{eq:lcb}. For the genetic algorithm, we propose Non-dominated Sorting Genetic Algorithm II (NSGA-II) \cite{deb2002nsgaii}, a widely used multi-objective evolutionary algorithm that has been applied successfully in the context of MOBO \cite{Belakaria2020usemo}. NSGA-II employs a combination of genetic operators, such as selection, crossover, and mutation, to evolve a population of candidate solutions. The algorithm utilizes a fast non-dominated sorting technique to rank individuals based on their dominance relationships, allowing for the identification of Pareto-optimal solutions in low computational time.

For solving the lower level, we propose the CTDE algorithm Multi-Agent Deep Deterministic Policy Gradient (MADDPG) \cite{lowe2017MADDPG}. In MADDPG, each agent has its actor-network that maps observations to actions and a centralized critic network that estimates the value function based on the joint actions of all agents. This centralized critic network allows agents to learn from the experiences of other agents, enabling them to better predict the actions that other robots will take \cite{he2020DMEDRL}. The actor-network is trained using the deterministic policy gradient algorithm, which updates the network parameters to maximize the expected return. This approach allows the agents to learn complex policies that can adapt to different exploration scenarios.

Algorithm~\ref{alg:BOFD} provides a functional algorithmic pseudo code for BOFD. The integration of multi-objective Bayesian optimization and MADDPG enables the exploration of the design space to identify promising fleet configurations, while the MARL algorithm enables the robots to learn and refine their policies based on the feedback received from the environment and the interaction with different agents. We highlight that the selection of AF, genetic algorithm and MARL algorithm can be replaced by any other method without loss of generality.
\begin{algorithm}[H]
\caption{BOFD Framework}\label{alg:BOFD}
\begin{algorithmic}
\STATE \textbf{Inputs:} Multi-objective function $[P(\mathbf{n}),C(\mathbf{n})]$, the number of total evaluations $T_{max}$, and the initial number of evaluations $t_0$, POMG $<N, S, O, \{\sigma_i\}, A, \{R_i\}, T, \gamma>$.
\STATE Initialize a $GP(0,\prod_{j=1}^{M} k_j(\mathbf{n},\mathbf{n}'))$ prior on $P(\mathbf{n})$.
\STATE Set $t \gets t_0$.
\STATE Evaluate $[P(\mathbf{n}),C(\mathbf{n})]$ at $t_0$ randomly chosen $\mathbf{n}$ to obtain a prior set $\{[P(\mathbf{n}_1),C(\mathbf{n}_1)],\cdots, [P(\mathbf{n}_{t_0}),C(\mathbf{n}_{t_0})]\}$.
\WHILE{$t \leq T_{max}$}
    \STATE Compute a GP posterior probability distribution of $P(\mathbf{n})$ conditioned on the prior set.
    \STATE Obtain AF score $LCB(\mathbf{n})$.
    \STATE Compute the population score as $\tilde{F}(\mathbf{n})= LCB(\mathbf{n}) + C(\mathbf{n})$.
    \STATE Calculate Pareto frontier $N_t$ of the cheap MO problem $\hat{F}(\mathbf{n})=[LCB(\mathbf{n}),C(\mathbf{n})]$ using NSGA-II initialized with the highest  $\tilde{F}(\mathbf{n})$ scores.
    \STATE Obtain the next fleet to be evaluated by selecting: 
    \STATE $\mathbf{n}' = \text{argmin}_{\mathbf{n}\in N_t} \tilde{F}(\mathbf{n})$.
    \STATE Run MADDPG to solve the POMG $<\mathbf{n}', S, O, \{\sigma_i\}, A, \{R_i\}, T, \gamma>$ and obtain the joint policy $\boldsymbol{\pi}$.
    \STATE Evaluate $[P(\mathbf{n}'),C(\mathbf{n}')]$ given $\boldsymbol{\pi}$ and append to the prior set.
    \STATE $t \gets  t+1$.
\ENDWHILE
\STATE \textbf{Return} $\mathbf{n}$ that minimizes $\mathcal{F}(\mathbf{n})$.
\end{algorithmic}
\end{algorithm}

\subsection{Theoretical Analysis}

In this section, we discuss the theoretical properties of the BOFD framework, specifically the upper bound for total regret and the effect of the bias initialization using a single-objective score for the initial population. 

We consider the multi-objective bound within the framework of LCB as an AF, as presented in \cite{Belakaria2020usemo}. This approach is an extension of the cumulative regret measure initially introduced for single-objective BO in \cite{srinivasetal2010lemma51}. Given the unique characteristics of the fleet design problem, our primary objective is to establish a sub-linear upper bound for cumulative regret. We generalize the multi-objective problem definition to:
\begin{equation}
min_{\mathbf{n}\in N} (F_1(\mathbf{n}), F_2(\mathbf{n})), 
\end{equation}
\noindent where $\mathbf{n} \in N$ is an $M$-dimensional vector containing the number of agents of each type in the fleet. $F_1(\mathbf{n})$ is an unknown function over the space $N$ and $F_2(\mathbf{n})$ a known cost function over the space $N$. 

\textbf{Theorem 1.} Let $\mathbf{n}^*$ be a solution in the Pareto set $N^*$ and $\mathbf{n}_t$ a solution in the Pareto set $N_t$ of the cheap multi-objective problem obtained at the $t^{th}$ iteration of BOFD. Let $R(\mathbf{n}^*) = ||R_1(\mathbf{n}^*), R_2(\mathbf{n}^*)||$, where $||.||$ is the norm of the vector, $R_1(\mathbf{n}^*)= \sum^{T_{max}}_{t=1} (F_1(\mathbf{n}_t)-F_1(\mathbf{n}^*))$ and $R_2(\mathbf{n}^*)= \sum^{T_{max}}_{t=1} (F_2(\mathbf{n}_t)-F_2(\mathbf{n}^*))$, then the  following holds with probability $1-\gamma$: 
\begin{equation}
R(\mathbf{n}^*) \leq \sqrt{k T_{max}\beta_{T_{max}}\gamma^i_{T_{max}}} + \max_a \mathbf{A} D\sqrt{T_{max}M},
\end{equation}

\noindent where $k$ is a constant and $\gamma^i_{T_{max}}$ is the maximum information gain about $F_1$ after $T_{max}$ iterations. The exploration factor is defined as $\beta=2\log(M\pi^2t^2/6\delta) $ where $\delta \in [0,1]$ is a confidence parameter. $\mathbf{A}$ is a $M$-dimensional vector $A=[a_1,...,a_M]^T$ where each entry is the acquisition cost of each robot type, and  $D$ is the size of the decision space. 

\textbf{Proof.} We aim to show a sub-linear regret for the multi-objective problem by proving a sub-linear upper bound for regret in each objective as in \cite{Belakaria2020usemo}.In this proof, we first derive the upper bound for $R_1(\mathbf{n}^*)$ and then $R_2(\mathbf{n}^*)$.  

Let $F_1$ be modeled using GP-LCB, then the cheap multi-objective optimization problem is defined as:
\begin{equation}
min_{\mathbf{n}\in N} (LCB_t(\mathbf{n}), F_2(\mathbf{n})), 
\end{equation}

\noindent then, the lower confidence bound (LCB) acquisition function for $F_1$ at any iteration $t$ can be defined as:

\begin{equation}
LCB_t(\mathbf{n}) = \mu_{t-1}(\mathbf{n}) - \beta^{1/2}_t \sigma_{t-1}(\mathbf{n}).
\end{equation}

According to Lemma 5.1 of \cite{srinivasetal2010lemma51},  the following inequality holds with probability $1-\gamma$:
\begin{equation}
|F_1(\mathbf{n}) - \mu_{t-1}(\mathbf{n})| \leq  \beta_t^{1/2}\sigma_{t-1}(\mathbf{n}).
\end{equation}

Under the assumption of optimality of $N_t$, either there exist a $\mathbf{n}_t\in N_t$ such that:
\begin{equation}
LCB_t(\mathbf{n}_t) \leq LCB_t(\mathbf{n}^*),
\end{equation}

\noindent or $\mathbf{n}^*$ is in the optimal Pareto set $N_t$ generated by multi-objective solver (i.e., $\mathbf{n}_t = n^*$).

Using lemma 1 of \cite{Belakaria2020usemo} for function $F_1$:
\begin{equation}
LCB_t(\mathbf{n}_t) \leq LCB_t(\mathbf{n}^*) \leq F_1(\mathbf{n}^*),
\end{equation}

\noindent then, according to Lemma 5.2 from \cite{srinivasetal2010lemma51}
\begin{equation}
R_1(\mathbf{n}^*) = F_1(\mathbf{n}_t) - F_1(\mathbf{n}^*) \leq F_1(\mathbf{n}_t)-LCB_t(\mathbf{n}_t),
\end{equation}
\begin{equation}
R_1(\mathbf{n}^*) \leq F_1(\mathbf{n}_t)-\mu_{t-1}(\mathbf{n}) + \beta^{1/2}_t \sigma_{t-1}(\mathbf{n}),
\end{equation}
\begin{equation}
R_1(\mathbf{n}^*)  \leq 2\beta^{1/2}_t \sigma_{t-1}(\mathbf{n}),
\end{equation}

\noindent considering Lemma 5.4 from \cite{srinivasetal2010lemma51}, $R_1(\mathbf{n}^*) \leq \sqrt{a T_{max}\beta_{T_{max}}\gamma^i_{T_{max}}} $ with probability $1-\gamma$.  

Now, we proceed to derive an upper bound for $R_2(\mathbf{n}^*)$. We proceed to show that $F_2$ is a Lipschitz continuous function in a bounded decision space. We aim to find an $L$ such that for any pair of vectors $\mathbf{n}_1$ and $\mathbf{n}_2$ in the $d$-dimensional space, the Lipschitz continuity condition holds:
\begin{equation}
|F_2(\mathbf{n}_1)-F_2(\mathbf{n}_2)| \leq L*||\mathbf{n}_1 - \mathbf{n}_2||,
\end{equation}

By the definition of cost function, we have $F_2(\mathbf{n})=\mathbf{A}*\mathbf{n}$. Then, 
\begin{equation}
|F_2(\mathbf{n}_1)-F_2(\mathbf{n}_2)| = |\mathbf{A}*\mathbf{n}_1 - \mathbf{A}*\mathbf{n}_2|,
\end{equation}

Since $\mathbf{A}$ is a $M$-dimensional positive constant, the Lipschitz constant $L$ depends on the components of $\mathbf{A}=[a_1,...,a_M]^T$ and we can set the upper bound to the maximum of the absolute values of the components of $\mathbf{A}$:
\begin{equation}
|\mathbf{A}*\mathbf{n}_1 - \mathbf{A}*\mathbf{n}_2| = |\mathbf{A}* (\mathbf{n}_1 - \mathbf{n}_2)|,
\end{equation}
\begin{equation}
|\mathbf{A}* (\mathbf{n}_1 - \mathbf{n}_2)| \leq \max_a \mathbf{A} ||\mathbf{n}_1 - \mathbf{n}_2||.
\end{equation}

The Lipschitz constant satisfies the Lipschitz continuity condition for all pairs of vectors $\mathbf{n}_1$ and $\mathbf{n}_2$ in the d-dimensional space.  Following \cite{Bubeck2015lipbound}, we can express the upper bound for $R_2(\mathbf{n}^*)$ as follows:
\begin{equation}
R_2(\mathbf{n}^*) \leq \mathbf{A} D\sqrt{T_{max}*M}.
\end{equation}

Therefore, the upper bound for total regret presented after $T_{max}$ iterations is:
\begin{equation}
R(\mathbf{n}^*)  \leq \sqrt{a T_{max}\beta_{T_{max}}\gamma^i_{T_{max}}} + \max_a \mathbf{A} D\sqrt{T_{max}M}.
\end{equation}

Theorem 1 states that as the number of iterations increases, the regret incurred converges to zero at a sub-linear rate. This property is highly desirable in optimization problems because it demonstrates that the approach becomes increasingly efficient and effective over time in terms of minimizing cumulative regret \cite{Srinivas2012regbound}. In practical applications like fleet design, where a large number of evaluations may be involved, sub-linear regret bounds are a key indicator of the algorithm's practical utility and scalability \cite{Shahriari2016reviewbo}. The sub-linear upper bound underscores the robustness and effectiveness of the approach in real-world scenarios, making it a fundamental metric for assessing the performance of the BOFD framework.

\section{Empirical results}

In this section, we conduct benchmark studies to assess the empirical performance of BOFD in synthetic and simulated problems. In the synthetic problem, we have access to the ground truth fleet performance, which allows us to compare the recommended fleet composition and the best fleet composition to obtain the regret of each benchmark method. In the simulated problem, we aim to train and deploy each fleet in different environments to assess their performance without knowing the optimal fleet. These experiments are particularly relevant as they address fundamental characteristics of robotics environments and multi-robot exploration tasks.

The benchmark study involves a variant of the BOFD framework utilizing USeMO \cite{Belakaria2020usemo} to address the fleet design problem at the upper level. To extend the single-objective bi-level optimization framework for reward design from \cite{shou2020reward} to discrete optimization, we propose three distinct approaches:
\begin{itemize}
    \item Naive BO (NBO): Implements a rounding operation in the recommended design to achieve a feasible discrete fleet.
    \item Discrete-BO (DBO) \cite{luong2019discretebo}: Optimizes the exploration factor of the acquisition function (AF) and the length scale of a covariance function to prevent premature repetition of designs.
    \item Probabilistic Reparametrization (PR) \cite{daulton2022pr}: Instead of directly optimizing the AF over discrete designs, it optimizes the expectation of the AF over a probability distribution defined by continuous parameters.
\end{itemize}

While both PR and DBO have demonstrated superior performance compared to NBO in various problems, there is a lack of benchmark studies comparing them. The implementation of PR is based on the official repository \footnote{\url{https://github.com/facebookresearch/bo_pr}}, and the implementation of DBO is derived from a public repository \footnote{\url{https://github.com/huuphuc2609/DiscreteBO/tree/master}}. NBO and USeMO are implemented based on the descriptions provided in the papers by \cite{shou2020reward} and \cite{Belakaria2020usemo}, respectively.

We consider two primary design features in the exploration robots: sensor range and field of view. Robots with long sensor ranges can perceive a larger portion of the environment, providing extensive information about distant objects and obstacles. In contrast, robots with short sensor ranges have limited perception, affecting their efficiency in exploring larger environments. Similarly, the field of view is a crucial factor. Robots with a wide field of view possess a comprehensive awareness of the environment, detecting objects and obstacles from various angles. On the other hand, robots with a narrower field of view may need to scan their surroundings more frequently to achieve similar coverage, impacting their overall efficiency.

We assume each robot can have either a long or short sensor range and a wide or narrow field of view, resulting in four distinct types of robots ($M=4$). The acquisition cost is assigned proportionally to the robot's capabilities, with fully-featured robots being the most expensive and low-featured robots the least expensive.

To define the search space, we consider the possibility of acquiring up to three robots per type, assuming the absence of robots in the environment is not a feasible solution. These assumptions result in a comprehensive search space comprising $|N|=255$ distinct fleet compositions. This search space complexity surpasses that of \cite{shou2020reward}, where the search space for their single-objective reward parameter is limited to 100 different values. This expanded search space adds an extra layer of complexity to the fleet design problem, posing a more challenging optimization task.

We conduct experiments using five distinct random seeds to evaluate the performance of BOFD across various scenarios, providing a comprehensive understanding of its capabilities. For each seed, a unique set of five initial random fleet compositions is generated, as detailed in Table \ref{tab:rpriors}. These initial random fleets serve as the prior information required by BO algorithms. This information is used to fit a Gaussian distribution, which is subsequently updated after each iteration based on the recommended fleets.  

\begin{table}[!t]
  \caption{Prior fleets per random seed. Each entry in the array shows the number of agents per type.}
  \label{tab:rpriors}
  \centering
  \begin{tabular}{|c|c|c|c|c|c|}
    \hline
   \textit{Prior}& \textit{Fleet 1} & \textit{Fleet 2} & \textit{Fleet 3} & \textit{Fleet 4} & \textit{ Fleet 5}\\
    \hline
    1 & [2,0,0,3] & [0,3,3,0] & [1,1,1,2] & [3,2,2,1] & [3,1,3,1]\\
    2 & [0,1,1,3] & [2,2,2,1] & [3,0,0,2] & [1,3,3,0] & [1,0,2,0]\\
    3 & [3,3,2,1] & [1,0,0,3] & [0,2,3,0] & [2,1,1,2] & [2,2,0,2] \\
    4 & [1,3,1,2] & [2,0,2,1] & [3,2,0,3] & [0,1,3,0] & [0,2,2,0] \\
    5 & [1,3,3,0] & [3,1,0,3] & [2,2,2,1]& [0,0,1,2] & [0,2,0,2] \\
    \hline
  \end{tabular}
\end{table}

In the subsequent sections, we provide detailed information on the implementation and present the results of experiments conducted in the synthetic problem and the simulated environments.

\subsection{Synthetic Experiments}\label{sec:synth}

The fleet design problem involving autonomous robots is characterized by an expensive-to-evaluate black-box function, with the primary bottleneck being the training and deployment of RL agents. In this set of experiments, we bypass the MARL algorithm by randomly assigning a performance value to each fleet, creating a known performance distribution. This approach enables us to benchmark the empirical regret of each method. Specifically, we compare the best-suggested fleet within a fixed budget of evaluations against the truly optimal fleet.

Regret is measured as the absolute difference between the best fleet found in each method and the optimal fleet under three different budgets: $T_{max}=10$, $T_{max}=15$, and $T_{max}=20$. Fleet performance is measured as the time required to complete exploration, randomly sampled from a uniform distribution $P(\mathbf{n}) \sim U(100, 750)$ for all $\mathbf{n} \in N$. The acquisition cost for full-feature robots is 150, for low-feature robots is 50, for robots with a long sensor range and narrow field of view is 100, and for robots with a short sensor range and wide field of view is 75, i.e. $\mathbf{A} = [150, 100, 75, 50]^T$. We report in Table \ref{tab:resultsS} the regret given by $\mathcal{F}(\mathbf{n})=P(\mathbf{n})+C(\mathbf{n})$ for each different random set of priors to allow comparison between single-objective and multi-objective methods. 

\begin{table}[!t]
  \caption{Regret in each set of random priors}
  \label{tab:resultsS}
  \centering
  \begin{tabular}{|p{1.3cm}|*{6}{p{0.7cm}|}}
    \hline
    \textit{$T_{\text{max}}=10$} & \textit{P1} & \textit{P2} & \textit{P3} & \textit{P4} & \textit{P5} & \textit{Avg.} \\
    \hline
    NBO & 112.9 & 39.03 & 35.38 & 233.23 & 5.79 & 85.27 \\
    DBO & 39.03 & 39.03 & 254.43 & 39.03 & 49.01 & 84.11 \\
    PR & 205.17 & 0.00 & 124.99 & 35.38 & 137.49 & 100.61 \\
    USeMO & 5.79 & 0.00 & 112.90 & 293.65 & 0.00 & 82.47 \\
    BOFD & 0.00 & 0.00 & 0.00 & 0.00 & 0.00 & 0.00 \\
    \hline
  \end{tabular}

  \begin{tabular}{|p{1.3cm}|*{6}{p{0.7cm}|}}
    \hline
    \textit{$T_{\text{max}}=15$} & \textit{P1} & \textit{P2} & \textit{P3} & \textit{P4} & \textit{P5} & \textit{Avg.} \\
    \hline
    NBO & 39.03 & 39.03 & 35.38 & 112.90 & 5.79 & 46.43 \\
    DBO & 39.03 & 39.03 & 205.17 & 39.03 & 49.01 & 74.25 \\
    PR & 205.17 & 0.00 & 124.99 & 35.38 & 137.49 & 100.61 \\
    USeMO & 0.00 & 0.00 & 0.00 & 112.90 & 0.00 & 22.58 \\
    BOFD & 0.00 & 0.00 & 0.00 & 0.00 & 0.00 & 0.00 \\
    \hline
  \end{tabular}

  \begin{tabular}{|p{1.3cm}|*{6}{p{0.7cm}|}}
    \hline
    \textit{$T_{\text{max}}=20$} & \textit{P1} & \textit{P2} & \textit{P3} & \textit{P4} & \textit{P5} & \textit{Avg.} \\
    \hline
    NBO & 39.03 & 39.03 & 35.38 & 112.90 & 5.79 & 46.43 \\
    DBO & 39.03 & 39.03 & 5.79 & 39.03 & 35.38 & 31.65 \\
    PR & 35.38 & 0.00 & 124.99 & 35.39 & 137.49 & 66.65 \\
    USeMO & 0.00 & 0.00 & 0.00 & 0.00 & 0.00 & 0.00 \\
    BOFD & 0.00 & 0.00 & 0.00 & 0.00 & 0.00 & 0.00 \\
    \hline
  \end{tabular}
\end{table}

We observe distinctive performance patterns among the methods with varying iteration budgets. BOFD stands out as the sole method capable of achieving the optimal solution, yielding zero regret across all priors. The earliest BOFD finds the optimal solution is at two iterations and the latest is at five iterations. With a budget of 10 iterations,  USeMO demonstrates commendable performance by achieving zero regret in two priors but faces challenges, notably in the fourth prior, leading to the highest regret in the synthetic experiments. Among single-objective methods, their average performance is comparable, with NBO and DBO mostly influenced by a single prior. Notably, PR achieves zero regret in one of the priors.

Expanding the budget to 15 iterations significantly benefits USeMO, allowing it to achieve zero regret in four priors and substantially reducing regret in the fourth prior. In contrast, single-objective methods, particularly NBO and DBO, show improvements in only a limited number of priors. PR struggles to enhance its performance, indicating that single-objective methods encounter difficulties in balancing the trade-off between performance and cost, resulting in the exploration of fleets with sub-optimal objective functions.

With a budget of $T_{\text{max}}=20$, USeMO manages to achieve zero regret across all priors. NBO fails to improve its regret in the additional iterations but still outperforms PR, which exhibits the highest average regret (66.65) and the maximum regret in the third prior  (124.99). DBO makes significant progress, reducing its regret by more than half and providing the lowest average regret among single-objective methods.

Overall, the results underscore the efficacy of multi-objective methods in swiftly reaching optimal fleet designs, with BOFD capitalizing MOBO better than USeMO, achieving the optimal solution in five or fewer iterations. Surprisingly, NBO demonstrates competitive performance compared to the specialized single-objective methods, outperforming PR in most scenarios and DBO when $T_{\text{max}}=15$. The promising performance of BOFD is further validated in a simulated environment in the subsequent section.

\subsection{Simulated Experiments}

\begin{figure}[!t]
\centering
\includegraphics[width=3.0in]{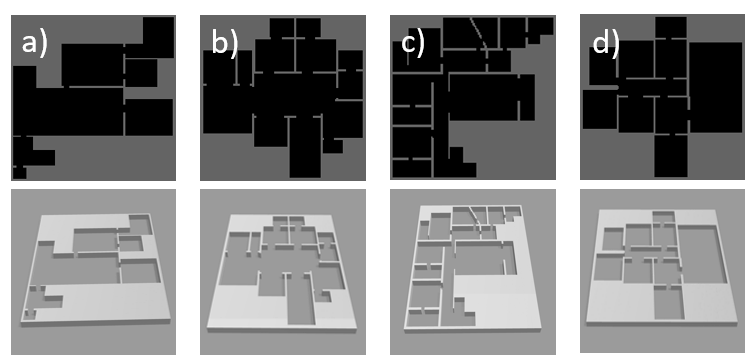}
\caption{Maps' layout for multi-robot exploration. The first row has the 2-dimensional version and the bottom row has the 3-dimensional representation.}
\label{fig:maps}
\end{figure}

In this section, we present benchmark studies conducted in simulated environments, featuring four distinct maps. Each map is generated using 2-dimensional layouts of four different sizes to introduce various levels of navigation complexity. Subsequently, we construct a 3-dimensional representation as illustrated in Figure \ref{fig:maps}. Map \#1 considers 120x120 grids with an acquisition cost of $\mathbf{A}_1=[400,320,375,300]^T$. Map \#2 spans 125x125 grids with an acquisition cost of $\mathbf{A}_2=[70,60,50,40]^T$. Map \#3 covers an area of 150x150 grids, accompanied by an acquisition cost of $\mathbf{A}_3=[150,120,140,95]^T$. Lastly, Map \#4 extends over 175x175 grids with an acquisition cost of $\mathbf{A}_4=[285,255,270,240]^T$. The deliberate diversity incorporated into the design of these environments leads to varying fleet performances and costs, thereby presenting unique optimization challenges for the fleet designer.

As in the synthetic experiments,  we aim to minimize total time to exploration while also minimizing the fleet acquisition cost, however, in this experimental setup, the performance of each fleet is the result of the joint policy learned by training the robots in the environments using MADDPG. The algorithmic implementation is based on the public repository\footnote{\url{https://github.com/hedingjie/DME-DRL}} of the work of \cite{he2020DMEDRL}. The Critic architecture considers two layers of Convolutional Neuronal Networks (CNN), a Long short-term memory (LSTM) network, and a fully connected layer. The input layer has $|S|+|A|$ neurons and the output layer is 1-dimensional. Each Actor follows a similar architecture with the difference that the input layer has a number of neurons equivalent to the observation space $|\sigma_i|$ of each agent $i$ to allow the decentralized execution. The main hyperparameters for MADDPG are shown in the Table  \ref{tab:hyperparameters}.

\begin{table}[!t]
  \caption{Hyperparameters for MADDPG}
  \label{tab:hyperparameters}
  \centering
  \begin{tabular}{|c|c|}
    \hline
     \textit{Hyperparamter} & \textit{Value}\\
    \hline
    Batch size & 60  \\
    Discount factor & 0.95  \\
    target network update factor & 0.001  \\
    Critic Learning Rate & 0.001 \\
    Actor Learning Rate & 0.0001 \\
    Timestep limit per episode & 50 \\
    Episodes & 5000 \\
    Field of view & \{5, 60\}\\
    Sensor range & \{10, 30\}\\
    \hline
  \end{tabular}
\end{table}

\begin{figure}[!t]
\centering
\includegraphics[width=2.5in]{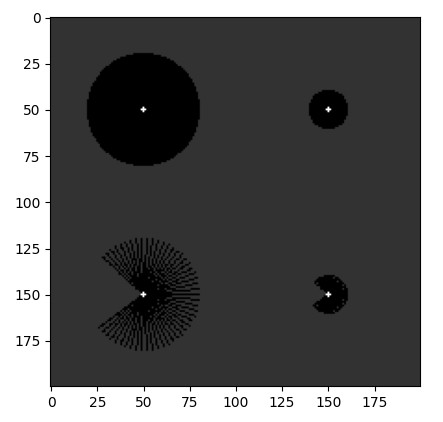}
\caption{Sensor capabilities for each of the four types of robots in a blank environment of 175x175 grids.}
\label{fig:sensors}
\end{figure}

The combination of field of view values and sensor range provides each type of robot with distinct exploration capabilities, shown in Figure \ref{fig:sensors}. Each fleet is trained to learn a joint policy using MADDPG in the 2-dimensional map. Then, the learned joint policy is deployed in the 3-dimensional environments implemented in a Gazebo multi-robot simulation, version 11.14, and the robots' models are built using ROS Noetic. We use TurtleBot 3 Burger as the base model for the exploration robots and we modify the Light Detection and Ranging (LiDAR) system parameters for range and field of view to obtain $M=4$ types of robots with different capabilities. 

\begin{figure}[!t]
\centering
\includegraphics[width=3.5in]{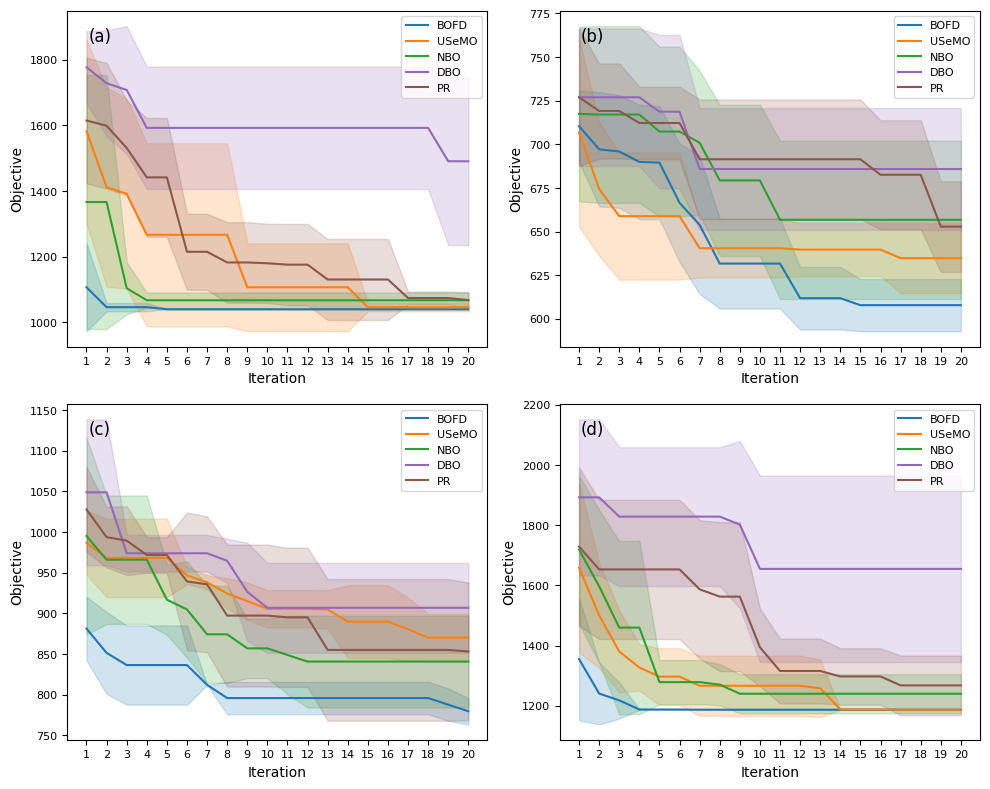}
\caption{Average objective value $\mathcal{F}(\mathbf{n})$ and standard deviation across five priors for (a) Map \#1, (b) Map \#2, (c) Map \#3 and (d) Map \#4.}
\label{fig:resultssim}
\end{figure}

The complexity of training multiple fleets makes it unfeasible to have access to the true performance distribution and compute the regret of each solution. Therefore we report the best performance obtained by each method measured as $\mathcal{F}(\mathbf{n})=P(\mathbf{n})+C(\mathbf{n})$ to allow comparison between single-objective and multi-objective methods. The comparison of the average and the standard deviation of best objective value $\mathcal{F}(\mathbf{n})$ across all prior sets in $T_{max}=20$ iterations are shown in Figure \ref{fig:resultssim}. 

In Map \#1, BOFD demonstrates remarkable efficiency by identifying promising fleet configurations within only two iterations across all priors, effectively minimizing the standard deviation resulting from diverse prior fleets within five iterations. Regardless of the prior set, BOFD consistently identifies the best solution $\mathbf{n} = [0,0,0,1]$, showcasing its robustness and effectiveness. USeMO initially exhibits high standard deviation, indicating a varied exploration across priors, yet it eventually converges to the best fleet in four out of five initial prior sets, closely mirroring PR's performance. NBO achieves convergence to the same fleet as BOFD in only two prior sets, but its tendency to converge early to sub-optimal designs impacts its overall performance. In contrast, DBO struggles to navigate the performance-cost trade-off effectively, failing to consistently improve fleet designs.

In Map \#2, BOFD stands out as the sole method to attain fleet $\mathbf{n} = [3,1,0,2]$, delivering the best objective value observed across all methods and priors. While USeMO initially discovers better fleets on average within the first seven iterations, BOFD surpasses it from the eighth iteration onwards. Among single-objective methods, PR converges to solutions with similar objective values as NBO, whereas DBO faces challenges in enhancing fleet designs after seven iterations. Overall, multi-objective methods exhibit superior performance compared to single-objective methods in this environment.

In Map \#3, BOFD exhibits a proactive approach by identifying promising fleet designs early in the search process and further improving them in subsequent iterations, ultimately securing the best design, $\mathbf{n} = [2,0,1,0]$, in four out of five prior sets. However, in the prior 3, BOFD achieves a slightly inferior fleet design. While NBO and PR manage to find the best fleet in two out of the five prior sets, their inability to replicate similar objective values across the remaining three sets affects their overall performance. Conversely, DBO and USeMO fail to reach the fleet offering the best objective value, resulting in worse average solutions compared to the other three methods.

In Map \#4, we observe a pattern similar to that of Map \#1, where BOFD consistently identifies promising fleet configurations within only two iterations across all priors. Moreover, it requires only five iterations to minimize the standard deviation resulting from diverse prior fleets, consistently achieving the best fleet $\mathbf{n} = [0,1,0,0]$. The noteworthy performance achieved in a few iterations holds particular significance in Map \#4, given the complexity of this environment. Notably, training the agents using MADDPG typically consumes an average of six hours per fleet configuration, highlighting the substantial time savings facilitated by the BOFD framework.

The results highlight the consistency of BOFD in navigating the exploration performance-cost trade-off, yielding fleet designs with minimal exploration times and costs. BOFD's performance aligns with the sub-linear regret outlined in Theorem 1, showcasing improved fleet designs as the iteration budget increases. Moreover, the tailor-made components of BOFD, such as the bias score $\tilde{F}$, the customized kernel, and the isolation of the uncertainty from the black-box objective, contribute to its superiority over other methods in addressing fleet design for autonomous robots.

We further analyze the trade-off between performance and cost in each map to evaluate whether the additional exploration time corresponds proportionally to a relevant reduction in acquisition costs of the recommended fleet. To accomplish this, we compare the fleet suggested by BOFD against the largest fleet (highest acquisition cost) and the fastest fleet (lowest exploration time), assuming equal importance for both metrics. The comparison for each map is presented in Table \ref{tab:bfleet}.

\begin{table}[!t]
  \caption{Fleet comparison in each map.}
  \label{tab:bfleet}
  \centering
  \begin{tabular}{|c|c|c|c|c|c|}
    \hline
    \textit{Map} & \textit{Fleet} & \textit{n} & \textit{Time} & \textit{Cost} & $\mathcal{F}$ \\
    \hline
    \multirow{3}{*}{1} & BOFD Fleet  & [0,0,0,1] & 740 & 300 & 1040\\
                       & Fastest Fleet  & [2,3,3,0]  & 81 & 2885 & 2966 \\
                       & Largest Fleet  & [3,3,3,3]  & 123 & 4185 & 4308\\
    \hline
    \multirow{3}{*}{2} & BOFD Fleet  & [3,1,0,2] &  231 & 350 & 581\\
                       & Fastest Fleet  & [3,3,3,3]  & 129 &  660  & 789 \\
                       & Largest Fleet  & [3,3,3,3]  & 129 & 660 & 789\\
    \hline
    \multirow{3}{*}{3} & BOFD Fleet  & [2,0,1,0] & 331 & 440 & 771\\
                       & Fastest Fleet  & [2,1,3,2]  & 165 & 1030 & 1195 \\
                       & Largest Fleet  & [3,3,3,3]  & 289 & 1515 & 1804\\
    \hline
    \multirow{3}{*}{4} & BOFD Fleet  & [0,1,0,0] & 932 & 255  & 1187\\
                       & Fastest Fleet  & [2,3,3,3]  & 225 & 2865 & 3090 \\
                       & Largest Fleet  & [3,3,3,3]  & 292 & 3150  & 3442\\
    \hline
  \end{tabular}
\end{table}

In Map \#1, BOFD trades 659 time units to achieve savings of 2,585 compared to the fastest fleet. Similarly, it trades 617 time units to realize savings of 3,885, indicating a trade-off where performance is decreased by fourfold to reduce costs by 14 times.

Moving to Map \#2, we observe it is the only map where the largest fleet is also the fastest fleet, which aligns with the analysis presented in \cite{yan2014team} where increasing the fleet size might lead to sub-optimal actions such as avoiding collisions. BOFD trades 102 time units for savings of 310. For Map \#3, BOFD trades 166 and 42 time units compared to the fastest and largest fleet, respectively, resulting in savings of 590 and 1,075. Lastly, in Map \#4, BOFD trades over 700 time units to save up to 2,895. The joint policies learned by each fleet in map \#4 are shown in the Supplementary Video (link: https://youtu.be/sS4iAXMNSqk). Across all scenarios, BOFD consistently provides fleet designs that offer greater cost-efficiency for the exploration task at hand.

\section{Conclusions}

We present a pioneering approach for algorithmic fleet design with autonomous robots, using MOBO to tackle the fleet design problem and a MARL algorithm to train the robots and evaluate fleet performance. Our method BOFD capitalizes on the inherent structure of the fleet design problem by integrating the known cost function and the Acquisition Function (AF) of performance to formulate a cheap MO problem. By obtaining the Pareto front, BOFD effectively navigates the trade-off between conflicting objectives.

The versatility of BOFD extends beyond exploration tasks, making it applicable to various multi-robot problems across diverse domains where fleet design involves balancing competing objectives. A key area for future research lies in addressing the primary bottleneck of fleet optimization—training agents using MARL algorithms. Introducing transfer learning mechanisms could offer a promising avenue to mitigate this bottleneck, enabling the reuse of learned knowledge across different fleet designs and reducing the computational burden of the optimization process.

\input{Main.bbl}



\end{document}